\documentclass{svproc}
\usepackage{url}
\usepackage{color}

\usepackage{times}
\usepackage{epsfig}
\usepackage{booktabs}

\usepackage{amsmath,amssymb,amsfonts}
\usepackage{algorithmic}
\usepackage{textcomp}

\usepackage{framed,multirow}
\usepackage{latexsym}

\usepackage{url}
\usepackage{multicol}
\usepackage{makecell}
\usepackage{blindtext}
\usepackage{cite}
\usepackage{tabularx}
\usepackage{subcaption}
\usepackage[pagebackref=true,breaklinks=true,letterpaper=true,colorlinks,bookmarks=false]{hyperref}

\begin{document}
\mainmatter              
\title{
JSTR: Judgment Improves Scene Text Recognition
}
\author{Masato Fujitake\inst{1}
}
\authorrunning{Masato Fujitake}
\tocauthor{Masato Fujitake}
\institute{
FA Research, 
Fast Accounting Co., Ltd.
, Japan
\\
\email{fujitake@fastaccounting.co.jp}
}

\maketitle              

\begin{abstract}
In this paper, we present a method for enhancing the accuracy of scene text recognition tasks by judging whether the image and text match each other. 
While previous studies focused on generating the recognition results from input images, our approach also considers the model's misrecognition results to understand its error tendencies, thus improving the text recognition pipeline. 
This method boosts text recognition accuracy by providing explicit feedback on the data that the model is likely to misrecognize by predicting correct or incorrect between the image and text.
The experimental results on publicly available datasets demonstrate that our proposed method outperforms the baseline and state-of-the-art methods in scene text recognition.
\keywords{
Scene text recognition, computer vision, machine learning}
\end{abstract}

\section{Introduction}
\label{sec:intro}
Text recognition in scene images is one of the active areas in computer vision~\cite{shi2016crnn, lyu2022maskocr}.
It is a fundamental and vital task in real-world applications such as scene understanding~\cite{zhang2022testr, fujitake2023a3s, fujitake2021tcbam, fujitake2024rl} and document understanding~\cite{yiheng2020layoutlm, fujitake2024layoutllm}.
Recognizing text in various fonts, colors, and shapes is challenging. 
Therefore, many methods have been proposed to address these issues.
Early work proposed a method that uses Convolutional Neural Networks (CNN) to utilize information from images and Recurrent Neural Networks (RNN) to recognize text sequences~\cite{shi2016crnn}. 
A preprocessing method to correct curved text images was also proposed to improve the accuracy~\cite{baek2021TRBA}. 
Recently, methods using powerful language models have been proposed for more robust recognition~\cite{bautista2022parseq, li2021trocr, fujitake2023dtrocr}. 
Although recognition performance has significantly improved with these methods, there are still issues with misrecognition in images that are difficult to recognize.

\begin{figure}[t]
    \centering
    \begin{subfigure}{1.\textwidth}
        \includegraphics[width=\linewidth]{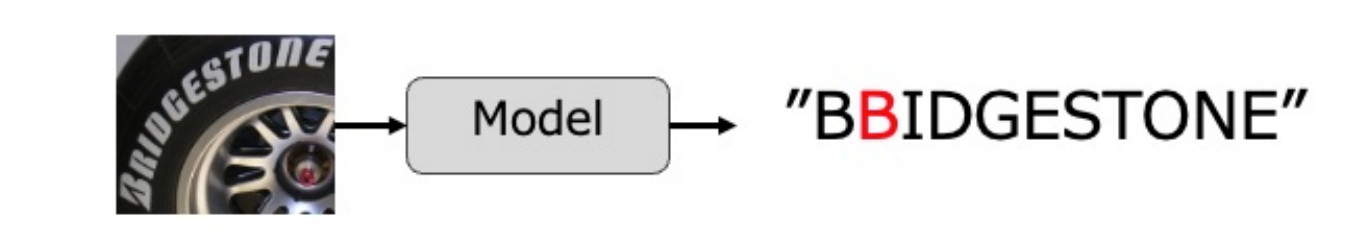}
        \caption{Typical approach of scene text recognition
        }
        \label{fig:recent_approach}
    \end{subfigure}\hfill \\
    \vspace*{1.00\baselineskip}
    \begin{subfigure}{1.\textwidth}
        \includegraphics[width=\linewidth]{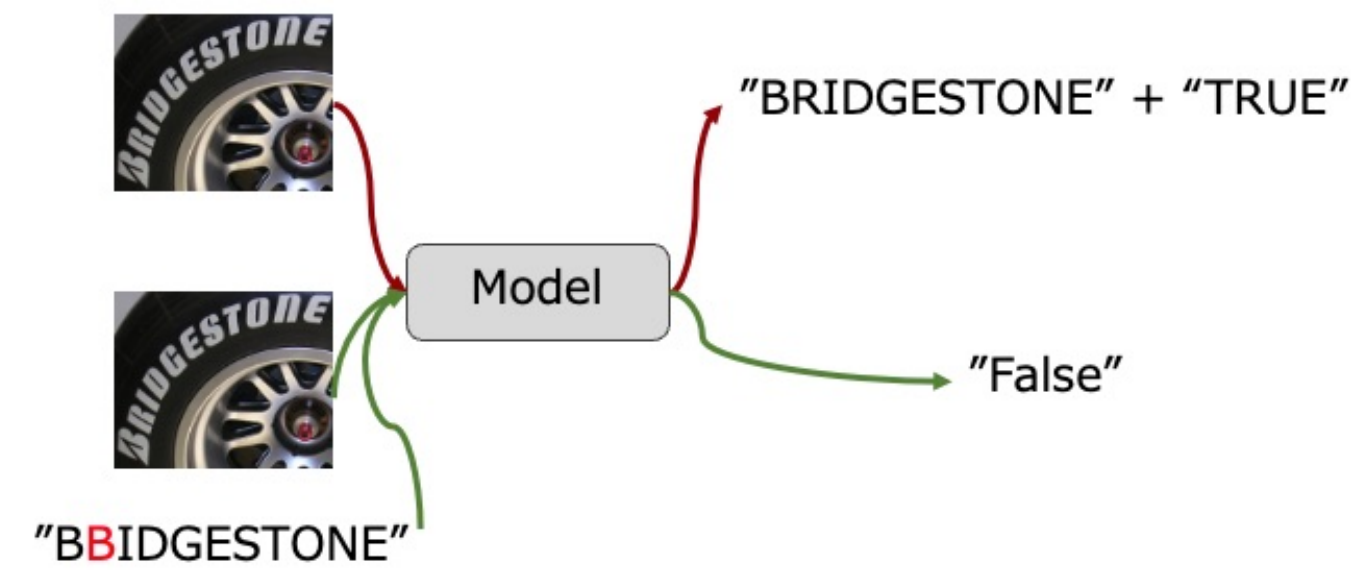}
        \caption{
        Our approach 
        }
        \label{fig:our_approach}
    \end{subfigure}\hfill
    \caption{
    \textbf{Differences in approaches between previous studies and the proposed method.}
 In the previous works, a model is formed to output recognition results from an image.
On the other hand, the proposed method performs modeling to judge whether the image and the text match each other in the same model, in addition to the modeling text recognition in the previous studies. 
If the image and text content match, the model judges the image as correct, and if not, the model makes the recognition wrong.
This improves recognition accuracy by strengthening the connection between the image and text.
    }
    \label{fig:abstract_approach}
\end{figure}

In this study, we introduce a novel framework, named JSTR, for recognizing text in scenes that concentrates on identifying data patterns that conventional text recognition models frequently misrecognize. 
This approach aims to enhance the ability to identify visual text by making explicit judgments about the error of its own results within the same text recognition model.
In the previous study, the model is trained to convert from image to text through the model, as shown in Figure ~\ref{fig:recent_approach}. 
The proposed method also learns to recognize image text as in previous studies.
However, it then learns to predict whether a pair of images and their recognition results are correct or not, using the recognition results. 
Identifying error patterns improves the model's discriminative ability for error-prone data. 
These are performed with a single model as shown in Figure ~\ref{fig:our_approach}, which outputs the recognition result and its correctness when an image is input and outputs its correctness when an image and a recognition result are input. 

The proposed method's ability to improve accuracy compared to the baseline has been demonstrated through experiments on publicly available benchmarks.
These results demonstrate the high effectiveness of JSTR, making it a beneficial choice for delivering superior performance.
\section{Related Works} There are two main approaches for scene text recognition: language-free and language-based. 
The language-free approach predicts text sequences directly from input images without any language requirements.
The main methods are based on Connectionist Temporal Classification  (CTC)~\cite{graves2006ctcloss} and segmentation-based methods. 
The CTC-based methods~\cite{hu2020gtc, shi2016crnn} combine sequence models such as RNNs with CNNs to extract visual features and predict text sequences with end-to-end training using CTC loss ~\cite{graves2006ctcloss}. 
Segmentation-based methods recognize text by segmenting and grouping them at the pixel level ~\cite{wan2020textscanner}. 
Recently, methods such as diffusion models have also been proposed for text recognition~\cite{fujitake2023diffusionstr}. 
However, since these methods do not use linguistic information but only image information, they tend to be vulnerable to noise, such as occlusion and distortion.

In order to increase robustness to these challenges, language-based approaches that apply constraints based on linguistic information have been studied in recent years~\cite{fang2021ABINet, fujitake2023dtrocr, li2021trocr}. 
Early studies proposed methods using N-grams, but more recently, methods using language models represented by RNNs and Transformer~\cite{vaswani2017transformer}~\cite{shi2018aster, fang2021ABINet } have been proposed. 
For example, ABINet obtains recognition results based on visual information and then inputs them into an external language model to obtain textual results with language information added. 
Then, it predicts a refined result by merging the two recognition results. 
PARSeq proposes Permuted Autoregressive Sequence Models that introduce an iterative improvement process ~\cite{bautista2022parseq}. 
TrOCR proposes to combine ViT~\cite{dosovitskiy2020vit}, which applies the Transformer to the visual information side, with a language model ~\cite{liu2019roberta}, which is pre-trained by Masked Language Modeling (MLM), to build an effective recognition model. 
Furthermore, in DTrOCR, ViT is eliminated, and a simple text recognition model is proposed based on the next word prediction problem, Generative Pre-Trained Transformer (GPT)~\cite{radford2019language}. 
Although these methods have significantly improved text recognition accuracy,  some misrecognition of hard-to-recognize characters still needs to be improved. 
Our proposed method is similar to previous studies in that it utilizes language models. 
However, the direct feedback approach of error tendencies generated by the models to the models is significantly different from previous studies that only learn to output text from images. 
\label{sec:relatedwork}
\section{Method} \label{sec:method}
In order to build the judgment mechanism of the model's output in the scene text recognition model, we have extended the language-based text recognition model DTrOCR~\cite{fujitake2023dtrocr} as a baseline.
In this section, we briefly review the baseline method and then propose our method, which includes true or false judgment for text recognition.

\begin{figure*}[h]
    \centering
    \includegraphics[width=1.0\textwidth]{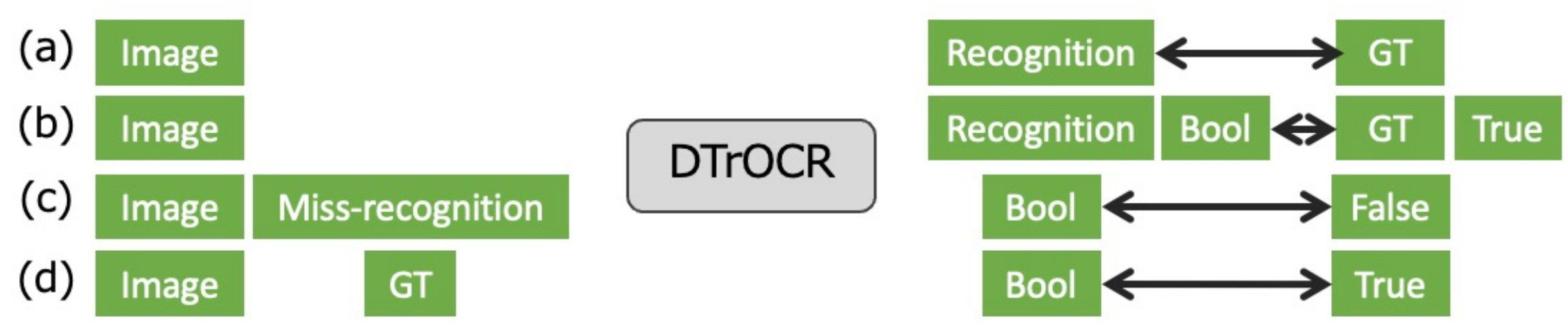}
    \caption{
\textbf{The flow of the data pipeline in the proposed method.}
The left column shows the input, model, output, and ground truth during training.
 Image, Recognition, GT, Bool, True, False, and Miss-recognition are the image, text recognition result, correct text of ground-truth, bool result of judgment, true, false, and misrecognition text, respectively. 
    }
\label{fig:proposed_method}
\end{figure*}

\subsection{DTrOCR}
DTrOCR~\cite{fujitake2023dtrocr} is a text recognition method based on the language model Generative Pre-Trained Transformer (GPT)~\cite{radford2019language}. 
GPT has been proposed in the natural language processing field and predicts sequential words for input text in an auto-regressive manner. 
GPT obtains linguistic ability by pre-training to predict the next word using a large corpus. 
It achieves high performance by utilizing this linguistic ability and fine-tuning the weights to the target task, such as summarization and translation. 

While GPT outputs a continuation of the input text based on its structure, DTrOCR inputs a patch sequence of images as shown in Figure~\ref{fig:proposed_method} (a), and outputs the recognition result as a text to follow it. 
The original GPT acquires only linguistic information, and the image and linguistic information are not connected. 
Therefore, DTrOCR solves this problem by generating an artificial dataset for text recognition using a large corpus and pre-training the language model GPT to link image and language. 
After pre-training, DTrOCR is fine-tuned using the target dataset, such as scene, printed, and handwritten text, to achieve more suitable text recognition. 

The algorithm first obtains a sequence of image features using only the Patch Embedding module of ViT~\cite{dosovitskiy2020vit}.
Then, the sequence is input to GPT, and the recognition result is output as a continuation of the sequence in an autoregressive manner by inputting a special token \texttt{[SEP]} that indicates the division of the image and the recognition result. 
The output is continued until the end of the recognition result, which is indicated by \texttt{[EOS]}. 
The recognition results are optimized using Cross Entropy loss with the corresponding ground truth text (GT).

\subsection{Proposed method}
The proposed method learns to judge whether a recognized text is correct after training the text recognition model.
The proposed method improves the discriminative ability of visual text recognition by judging correct / error recognition on an image-text pair dataset. 
This allows for more accurate identification even in ambiguous text images and is prone to errors.

The proposed method consists of two major training steps: text recognition and correct/incorrect judgment. 
First, as a text recognition training step, we train the baseline method, DTrOCR, as in the previous work. 
In this process, the model takes an image as input and outputs the recognition results as shown in Figure ~\ref{fig:proposed_method} (a). 
Next, judgment training is performed in addition to text recognition. 
The prediction of correct/false decisions is extended from the output of DTrOCR by connecting the text recognition results and the true/false decisions (True, False) with the special token for the separator \texttt{[SEP]}. 
In this way, the model interacts with the image sequence, the recognized text, and the set of true/false decisions as a single sequence. 
Therefore, in this process, the model is trained with the three kinds of input-output patterns shown in Figure~\ref{fig:proposed_method} (b-d). 
In pattern (b), similar to (a), the model takes an image as input and, in addition to outputting the usual recognition result, outputs a Bool to indicate whether the output is correct or not. 
Then, the pair of GT text and True is used as the ground truth for optimization. 
Patterns (c) and (d) are input/output patterns that learn only correct and incorrect decisions; they take an image and a pair of misrecognition results or GT and predict whether they are matched or not.

To create misrecognized results for the judgment data, the inference is performed on the training dataset using the trained text recognition model, and the result that does not match GT text is considered a false recognition result. 
We use the misrecognized text and its associated image as the false data of the judgment dataset.
In addition, the true data result pair for the judgment is generated using the correct text and its image for the misrecognized result.
Therefore, we use the correct and misrecognized text pair data for each misidentified image to generate the training data for the true/false judgment.
\section{Experiments} \label{sec:experiments}

\subsection{Dataset and Evaluation}
In order to make a fair comparison, we experimented following the setting of the existing work~\cite{fujitake2023dtrocr}. 
For evaluation, we use word-level accuracy on the six benchmark datasets.
The accuracy is considered correct if and only if the predictions match the characters in all positions. 
Following the previous study ~\cite{fujitake2023dtrocr}, we report the average scores of the four experiments. 
Six standard benchmarks were used for the evaluation. ICDAR 2013 (IC13)~\cite{karatzas2013icdar}, ICDAR 2015 (IC15)~\cite{karatzas2015icdar}, IIIT 5KWords (IIIT)~\cite{mishra2012iiit}, Street View Text (SVT)~\cite{wang2011svt}, Street View Text-Perspective (SVTP)~\cite{phan2013svtp}, CUTE80 (CUTE)~\cite{risnumawan2014cute80}. 

Two datasets are used for training: a synthetic dataset and a real image dataset. 
For synthetic dataset, the models are trained on two synthetic datasets, MJSynth (MJ)~\cite{jaderberg2014MJSynth, jaderberg2016MJSynth} and SynthText (ST)~\cite{gupta2016synthtext}. 

In recent years, there has been a trend in previous studies to evaluate text recognition using real data because it is more suitable for real-world settings.
In this study, we also fine-tuned the recognition model using real data to confirm its effectiveness. 
Specifically, by the previous studies, we fine-tuned the following data: COCO-Text~\cite{veit2016coco}, RCTW~\cite{shi2017icdar2017}, Uber-Text~\cite{zhang2017uber}, ArT~\cite{chng2019icdar2019}, LSVT~\cite{sun2019icdar}, MLT19~\cite{nayef2019icdarmlt}, ReCTS~\cite{zhang2019icdar}.

\subsection{Implementation Details}
We follow the previous work~\cite{fujitake2023dtrocr} exactly to reproduce the model.
The average accuracy on the benchmark was 0.1 points higher than that of the previous study. 
We used ADAMW~\cite{loshchilov2018adamw} for the second training step, which includes the judgment process, warmed up linearly to a learning rate of $2^{-6}$, and decreased to 0 according to the cosine decay. 
The number of training epochs is 10, and the number of warm-up epochs is two. 
The training data for correctness judgment were created based on the proposed method section and were randomly mixed with the text recognition dataset to construct the whole dataset. 
All experiments were conducted on four Nvidia A100 GPUs with mixed precision using PyTorch.

\begin{figure}[htp]
	\centering
	\includegraphics[width=1.00\columnwidth, keepaspectratio]{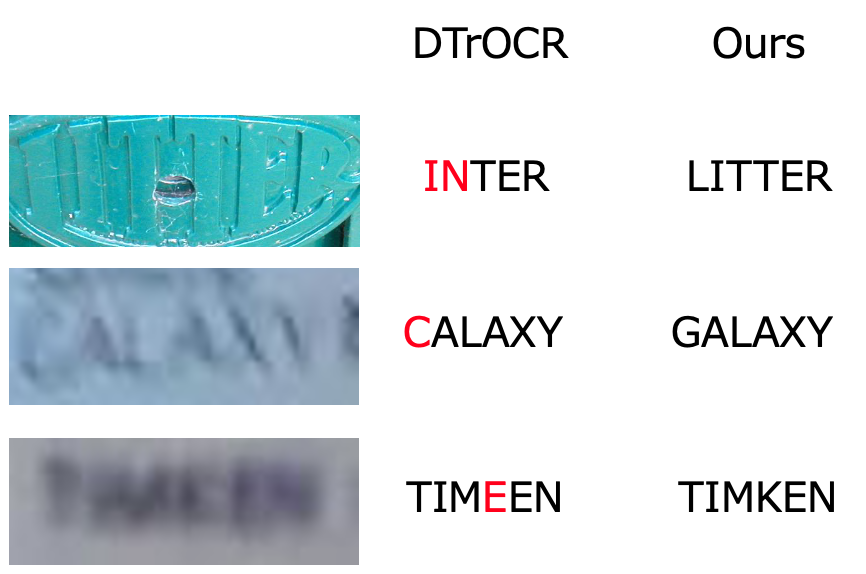}
	\caption{
\textbf{Comparison of recognition results in baseline (DTrOCR~\cite{fujitake2023dtrocr}) and proposed methods.}
The leftmost column is the input image, and each column shows the recognition results for each method.
Black characters indicate cases that match the ground truth, and red ones indicate misrecognition cases.
Comparison with the baseline confirms that the proposed method is robust in hard-to-read cases because it learns misrecognition tendencies.
	}
	\label{fig:visualized_comparison}
\end{figure}

\subsection{Main Result}
Table~\ref{tab:method_overall_result_english} shows the results of the state-of-the-art methods, the baseline model DTrOCR and our proposed method. 
Each method is listed in a separate section depending on the difference in the training dataset.
Compared to the baseline, our method improves the accuracy on all benchmarks in the synthetic dataset, showing that the correct/incorrect judgment is practical for text recognition. 
We can also confirm that our method reaches competitive accuracy against other powerful methods. 

When the model is trained on real-world images, the proposed method improves the recognition accuracy more, as in the previous works~\cite{fang2021ABINet, fujitake2023dtrocr}.
Among the models trained on real-world images, the proposed method achieves better accuracy than the baseline model.
It is also confirmed to have higher recognition accuracy than previous studies.
The effectiveness of the proposed method has been confirmed through experiments. 
The results show that the final recognition accuracy improves by learning the error tendency of the model through correct and incorrect judgments.

\subsection{Detailed Analysis}
Table~\ref{tab:ablation_model} displays the results of ablation studies conducted on synthetic datasets in a controlled environment. 
The average score for each benchmark is presented.
The baseline score was 96.0 \%, while the proposed method achieved a score of 97.3 \%.
The number of training iterations is increased since the proposed method is trained in two steps. 
In order to account for this, we trained the baseline model with an extended number of training iterations.
In this case, no significant difference from the baseline model is observed. 
No significant effect on accuracy is observed when including only the True pair data of the judgment, which are the same input/output pattern as the proposed method. 
The significant increase in accuracy after the judgment training with the true and false pairs confirms the effectiveness of error recognition.

Several examples of recognition results by the baseline model and the proposed method are presented in Figure~\ref{fig:visualized_comparison}.
Our proposed method performs well for unclear images and hard-to-read text images. 
It has been observed that the proposed method can reduce misrecognition by learning its error tendency.

\begin{table*}[tb]

\centering
   \caption{
\textbf{
Word accuracy on English scene text recognition benchmark datasets with 36 characters. 
``Synth'' and ``Real'' refer to synthetic and real training datasets, respectively.
}
}
\label{tab:method_overall_result_english}
\begin{tabular}{cccccccccc}
\toprule
\multirow{3}{*}{Method} &\multirow{3}{*}{Training data} & \multicolumn{8}{c}{Test datasets and \# of samples}                                                                                                                                                                \\ \cline{3-10} 
&                         & IIIT5k                   & SVT                      & \multicolumn{2}{c}{IC13}                            & \multicolumn{2}{c}{IC15}                            & SVTP                     & CUTE                     \\ 
&                         & 3,000                    & 647                      & 857                      & 1,015                    & 1,811                    & 2,077                    & 645                      & 288                      \\ 
\midrule
CRNN~\cite{shi2016crnn} & Synth &81.8 & 80.1 & 89.4 & 88.4 & 65.3 & 60.4 & 65.9 & 61.5 \\ 
TRBA~\cite{baek2021TRBA} & Synth &92.1 & 88.9 & $-$ & 93.1 & $-$ & 74.7 & 79.5 & 78.2 \\
ABINet~\cite{fang2021ABINet} & Synth & 96.2 & 93.5 & 97.4 & $-$ & 86.0 & $-$ & 89.3 & 89.2 \\

DiffusionSTR~\cite{fujitake2023diffusionstr} & Synth & 97.3 & 93.6 & 97.1 & 96.4 & 86.0 & 82.2 & 89.2 & 92.5 \\

$\textrm{TrOCR}_{\rm BASE}$~\cite{li2021trocr} & Synth & 90.1 & 91.0 & 97.3 & 96.3 & 81.1 & 75.0 & 90.7 & 86.8  \\
$\textrm{TrOCR}_{\rm LARGE}$~\cite{li2021trocr} & Synth & 91.0 & 93.2 & 98.3 & 97.0 & 84.0 & 78.0 & 91.0 & 89.6  \\

PARSeq~\cite{bautista2022parseq} & Synth& 97.0 & 93.6 & 97.0 & 96.2 & 86.5 & 82.9 & 88.9 & 92.2 \\
$\textrm{MaskOCR}_{\rm BASE}$~\cite{lyu2022maskocr} & Synth & 95.8 & 94.7 & 98.1 & $-$ & 87.3 & $-$ & 89.9 & 89.2 \\
$\textrm{MaskOCR}_{\rm LARGE}$~\cite{lyu2022maskocr} & Synth& 96.5 & 94.1 & 97.8 & $-$ & 88.7 & $-$ & 90.2 & 92.7 \\

DTrOCR~\cite{fujitake2023dtrocr} & Synth & 98.4 & 96.9 & 98.8 & 97.8 & 92.3 & 90.4 & 95.0 & 97.6 \\
Ours &Synth & \textbf{99.0} & \textbf{98.4} & \textbf{99.2} & \textbf{98.7}  & \textbf{94.8} & \textbf{93.2} & \textbf{96.6} & \textbf{98.5} \\

\hline
\hline
CRNN~\cite{shi2016crnn} & Real &94.6 & 90.7 & 94.1 & 94.5 & 82.0 & 78.5 & 80.6 & 89.1 \\ 

TRBA~\cite{baek2021TRBA} & Real &98.6 & 97.0 & 97.6 & 97.6 & 89.8 & 88.7 & 93.7 & 97.7 \\
ABINet~\cite{fang2021ABINet} & Real & 98.6 & 97.8 & 98.0 & 98.0 & 90.2 & 88.5 & 93.9 & 97.7 \\

PARSeq~\cite{bautista2022parseq} & Real& 99.1 & 97.9 & 98.3 & 98.4 & 90.7 & 89.6 & 95.7 & 98.3 \\
DTrOCR~\cite{fujitake2023dtrocr} & Real & 99.6 & 98.9 & 99.1 & 99.4 & 93.5 & 93.2 & 98.6 & 99.1 \\
Ours &Real & \textbf{99.8} & \textbf{99.5} & \textbf{99.4} & \textbf{99.6}  & \textbf{96.1} & \textbf{95.4} & \textbf{99.3} & \textbf{99.8} \\
%
%
\bottomrule

\end{tabular}
\end{table*}

\begin{table}[!t]
  \caption{
\textbf{
Ablation study.
}
}
\label{tab:ablation_model}
  \centering
\resizebox{0.6\columnwidth}{!}{
  \begin{tabular}{lc}
    \toprule
    Method & Average  \\
    \midrule
    Reproduced DTrOCR (baseline) & 96.0 \\
    Baseline w/ same iter  & 95.9 \\
    Proposed method w/o only-true &  96.1 \\
    Proposed method  & 97.3 \\ 
  \bottomrule
\end{tabular}
}
\end{table}

\section{Conclusion}\label{sec:conclusion}
This work presented a new framework called JSTR for scene text recognition tasks. 
Unlike other methods that only focus on recognizing text, JSTR also learns to distinguish between correct and incorrect recognitions by pairing an image with its recognition result. 
This judgment task helps to enhance the matching between the image and text, leading to more reliable text recognition and improved accuracy compared to previous methods. 
Experimental results have confirmed the effectiveness of JSTR on real datasets.

\clearpage
\bibliographystyle{unsrt}
\bibliography{article}

\end{document}